# Usage of Decision Support Systems for Conflicts Modelling during Information Operations Recognition

Oleh Andriichuk[1,2], Vitaliy Tsyganok[1], Dmitry Lande[1,2], Oleg Chertov[2], Yaroslava Porplenko[1]

[1]Institute for Information Recording of National Academy of Sciences of Ukraine, Kyiv, Ukraine
[2]National Technical University of Ukraine "Igor Sikorsky Kyiv Polytechnic Institute", Kyiv, Ukraine

`andriichuk@ipri.kiev.ua`, `tsyganok@ipri.kiev.ua`, `dwlande@gmail.com`, `chertov@i.ua`, `daliss@ukr.net`

**Abstract.** Application of decision support systems for conflict modeling in information operations recognition is presented. An information operation is considered as a complex weakly structured system. The model of conflict between two subjects is proposed based on the second-order rank reflexive model. The method is described for construction of the design pattern for knowledge bases of decision support systems. In the talk, the methodology is proposed for using of decision support systems for modeling of conflicts in information operations recognition based on the use of expert knowledge and content monitoring.

## 1 Introduction

Data sources make an essential effect on the people. During last years, it has become evident that mass media may be efficiently used for spreading of misinformation [1]. In addition, the social experiments show that many people believe to unconfirmed news and continue to spread it. For example, in [2] the review is presented of known false opinions and misinformation in the American society. In [3] the social experiments are described in which the susceptibility of people to political rumors about the health care reform approved by the USA Congress in 2010 is studied. In dependence on the way of information presentation, 17-20% of the experiment participants believed in the rumors, 24-35% of participants did not have a definite opinion, and 45-58% of the questioned rejected them.

By information operation (IO) [4] the complex of information events (news posts in Internet and articles in newspapers, news on radio and TV, comments and likes in social networks, forums, etc.) is meant which targets the modification of the social opinion about the certain subject (person, organization, institution, country, etc.). IO can consist of the system of information attacks and can take a long time. For example, distribution of rumors about problems in a bank can provoke its depositors to take back the deposits and in turn can lead to its bankruptcy. In general, to IO the misinformation events are related.

In [4] it is shown that IO is related to so-called weakly structured subject domains since they possess some specific for such areas features: uniqueness, impossibility of function target formalization, and, as a consequence, impossibility of construction of an analytical model, dynamicity, incomplete description, presence of the human factor, absence of standards. Such subject areas are treated using the expert decision support systems (DSS) [5].

By a conflict, a subject opposition in a search for a scarce resource in the media is meant. In the case of IO, the conflict is the effect to the audience.

## 2 Conflicts Modelling Approaches

There exist a number of approaches and models from theory of games for conflict modelling. The developed mathematical models of the conflict help their participants to build the optimum strategy. This is also related to the theory of games suggested by O. Morgenstern and J. von Neumann [6]. This theory was developed in many papers in which different aspects of the opposition process were considered [7, 8]. The Nash equilibrium [9] as a method of resolving non-cooperative games should also be mentioned. In addition, the papers appeared, in which the opposition process is considered not as the measure for the victory of one player over another, but the tool to determine the ways of interaction of parties [10]. Shelling was one of the first who applied the theory of games to international relations considering the armament race in [10]. In this paper, he considered the long-term conflicts and concluded that establishment of continuous friendly relations between the parties can lead to higher profit (even with account of higher losses during this period), than in short-term relations.

In classical works on game theory, the player's profit is determined by the constant predetermined payment matrix, which in many cases is quite difficult to obtain. At present, the method of target dynamic estimation of alternatives [11], based on the application of a hierarchy of goals, is widely used for decision-making in complex target programs. This method allows to calculate the effectiveness of each alternative (in this case – a possible player's turn). Analytical expression to describe the winning players in most cases is difficult (sometimes impossible), and the use of a hierarchy of goals is very convenient for describing the preferences of players. It is reasonable to use this model for conflicts modelling, taking into account conflicts subjects's reflexion, and departure from a certain sequence of steps and a transition to a scenario in which each of the subjects performs a complex of actions in dynamics. Reflexive models [12, 13] allow also taking into account the subjectivity of the opposite parties and the presence of compromises in some items.

## 3 Reflexive Model of Subjects in Conflict

The complete model describing the readiness of the subject with reflexion to accomplish some action (model of reflexion subject choice based on the second-order rank reflexive model [12]) is described by the following function (1):

$$A = (a_3 \rightarrow a_2) \rightarrow a_1 \qquad (1)$$

where $A$ is the subject's choice readiness; $a_1$ is the influence of the environment on the subject; $a_2$ is the psychological setting of the subject (the influence of the environment expected by the subject); $a_3$ – the subject's intensions.

The equation describing the self-esteem of the subject in the situation is as follows (2):

$$A_1 = a_3 \rightarrow a_2 \qquad (2)$$

where $A_1$ is the self-esteem of the subject in the situation.

The self-esteem of the subject in the situation means the appearance of "self-image" in "self-image", when the subject estimates his own image of the situation, his imagination about himself and his intentions.

Let us consider the general scheme of interaction of two subjects A and B in conflict conditions, being subjected to the same influence of the external environment [12, 13]. Subject A supposes that his counterpart B also possess the reflexion, i.e. has his own imaginations about the environment effect, his plans and wishes in the situation. In this case subject A in some manner is interpreting own relations with subject B and his ideas about these relations. Then the subject A choice readiness is described by the following function (3):

$$A = (a_3 \,\&\, b_3 \rightarrow a_2) \vee (a_4 \,\&\, b_4 \rightarrow b_2) \rightarrow a_1 \qquad (3)$$

where $A$ is the subject A choice readiness; $a_1$ is the influence of the environment on both subjects; $a_2$ is expected by subject A influence of the environment; $b_2$ is expected by subject B influence of the environment from the point of view of subject A; $a_3$ – intentions of subject A; $b_3$ – intentions of subject B from the point of view of subject A; $a_4$ is the impression of subject A of how subject B imagines the intentions of subject A; $b_4$ is the impression of subject A of how subject B imagines his own intentions.

The expression which describes the self-esteem of subject A in the conflict situation with subject B is as follows (4):

$$A_1 = (a_3 \,\&\, b_3 \rightarrow a_2) \vee (a_4 \,\&\, b_4 \rightarrow b_2) \qquad (4)$$

where $A_1$ is the self-esteem of subject A in the conflict situation with subject B.

The subject B choice readiness is described by the following function (5):

$$B = (c_3 \ \& \ d_3 \rightarrow c_2) \lor (c_4 \ \& \ d_4 \rightarrow d_2) \rightarrow a_1 \qquad (5)$$

where $B$ is the subject B choice readiness; $a_1$ is the influence of the environment on both subjects; $c_2$ is the expected by subject B influence of the environment; $d_2$ is the expected by subject A influence of the environment from the point of view of B; $c_3$ – intensions of subject B; $d_3$ – intensions of subject A from the point of view of subject B; $c_4$ is the impression of subject B about how subject A imagines the intensions of subject B; $d_4$ is the impression of subject B about how subject A imagines his own intentions.

The expression which describes the self-esteem of subject B in the conflict situation with subject A is as follows (6):

$$B_1 = (c_3 \ \& \ d_3 \rightarrow c_2) \lor (c_4 \ \& \ d_4 \rightarrow d_2) \qquad (6)$$

where $B_1$ is the self-esteem of subject B in the conflict situation with subject A.

Let us consider how the described above model of reflexive behavior of subjects in the conflict can be used for construction of the knowledge base (KB) for DSS.

## 4 Reflexive Model of Conflict in Knowledge Bases of Decision Support Systems

The result of the conflict of subjects will depend on the degree of goal achievement for each subject of the conflict. The winner will be the subject with higher value of the goal achievement degree. Thus, the model of the subject area of the conflict should be constructed in DSS KB in order to calculate the respective values of goal achievement degrees.

Based on the features of target decomposition in construction of a goal hierarchy graph and applying the method of target dynamic estimation of alternatives [11] one can assign some logical operations to DSS KB objects and to their relations. By means of DSS, in framework of which the above-mentioned method has been implemented, it is possible to model logical "or" ($\lor$) as subgoals of one goal, logical negation as a negative influence of the respective goal, "XOR" as groups of goal compatibility.

Analyzing the presented above reflexive model of the conflict of two subjects we can suggest the following design pattern for DSS KB (Fig. 1). Black solid arrows indicate the positive influence of goals and the dashed red arrows indicate the negative influence. Titles of the goals correspond to designations in the above equations.

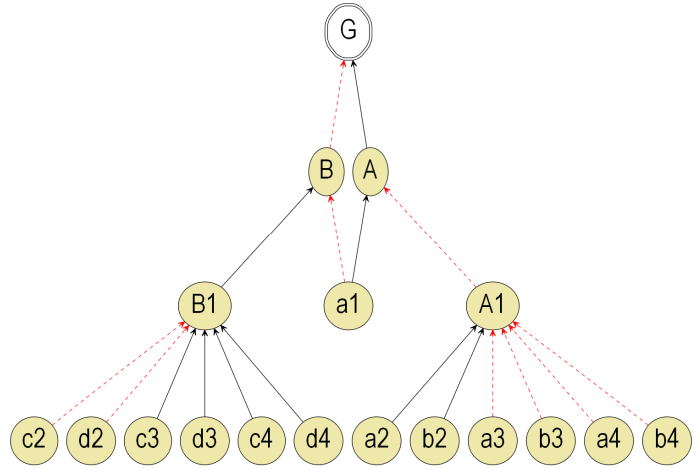

**Fig. 1.** Decision support system's knowledge bases design pattern for the conflict of two subjects.

In construction of this design pattern for DSS KB (Fig.1) the above-mentioned features of modelling of logical operations has been taken into account. For assignment of the logical equation with objects and relations in DSS KB the functions describing the choice readiness of subjects A and B were transformed as follows in (7) and (8):

$$A = \neg A_1 \vee a_1 \qquad (7)$$

$$B = \neg B_1 \vee a_1 \qquad (8)$$

For assignment of the logical equation with objects and relations in DSS KB the functions describing the self-esteems of the subjects in the conflict situation were transformed as follows in (9) and (10):

$$A_1 = a_2 \vee b_2 \vee \neg a_3 \vee \neg b_3 \vee \neg a_4 \vee \neg b_4 \qquad (9)$$

$$B_1 = c_2 \vee d_2 \vee \neg c_3 \vee \neg d_3 \vee \neg c_4 \vee \neg d_4 \qquad (10)$$

Since the winner will be the subject with higher value of the goal achievement degree, let us assume that the main goal of the hierarchy G (Fig.1) is affected positively by the goal of subject A, and negatively by the goal of subject B. Then if the value of the achievement degree of the main goal G is above zero, then subject A wins in the conflict, if this value is less than zero, then subject B is the winner, and in the case of zero value there is a draw.

DSS allows also to calculate separately the goal achievement degrees for subject A and subject B.

Obtained in such a way DSS KB design pattern should be complemented to the full range KB using the methods of DSS KB construction in identifying of information operations [14, 15]. Supplementation of KB with the use of expert knowledge only, as is described in [15] requires the group of experts. The work of experts is rather expensive and requires much time. Also for small expert groups the competency of experts should be taken into account [16], leading to additional time expenses for the search of extra information and for its processing. Therefore, expert information should be used moderately in the process of DSS KB construction for IO recognition.

## 5 Methodology for Conflicts Modelling by Decision Support Systems in Information Operations Recognition

Thus, the essence of the suggested method for modelling of conflicts using DSS during IO recognition is as follows:

1) The preliminary study of the IO object is carried out, the subjects of the conflict are determined, together with their goals, related to the subjects of the conflict, persons, organizations, companies. In the process of informational and analytic research a common problem, often faced by analysts is compiling a ranking of a set of objects or alternatives (products, electoral candidates, political parties etc) according to some criteria. For this task should apply special approaches for considering importance of information sources during aggregation of alternative rankings [17].

2) The respective design pattern for DSS KB is selected and modified if necessary. In modification of the design, one should take into account the features of modelling of logical operations in DSS KB.

3) The selected design pattern for DSS KB is complemented to full-range KB. The group expertise on determination and decomposition of IO goals is carried out. Thus, the decomposition of IO as of complex weakly structured system is taking place. For this purpose, the system for distributed acquisition and processing of expert information (SDAPEI) is used.

4) The respective KB is complemented using DSS tools taking into account the results of the group expertise, carried out by means of SDAPEI, and available objective information. For clarification of queries to content monitoring systems (CMS) and for complementation of DSS KB with lacking objects and links the keyword network of the subject area [14] of respective IO is used.

5) Using CMS tools the analysis of dynamics of the thematic data stream is carried out. DSS KB is complemented with partial influence coefficients.

6) Using DSS tools based on the constructed KB the recommendations are calculated.

Recommendations (in the form of dynamic estimations of efficiency of topics related to the IO object and values of the goals achievement degrees of subjects of the conflict), obtained by the above described technique, are used for the estimation of the IO-related damage [15], and for organization of the information counter-actions with account of information sources, as well as to predict result of the conflict.

The suggested method of modelling the opposition of two subjects can be used for IO recognition for modelling of confrontation of lobbyists and their opponents, for example, in such an event as Brexit [18].

## 6 Conclusions

In the talk, the advantages of the use of DSS in modelling of conflicts during the information operations recognition are substantiated. An information operation is considered as a complex weakly structured system.

The model of the conflict between two subjects is presented based on the second-order rank reflexive model.

The method is described enabling the construction of DSS's knowledge bases based on the model of the conflict between two subjects.

The method of application of the DSS for modelling the conflicts in information operation recognition is suggested. The ranges of applicability of this method are suggested.

## Acknowledgment

This study is funded by the NATO SPS Project CyRADARS (Cyber Rapid Analysis for Defense Awareness of Real-time Situation), Project SPS G5286.